\documentclass[lettersize,journal]{IEEEtran}
\usepackage[utf8]{inputenc}  
\usepackage{amsmath,amsfonts}
\usepackage{algorithmic}
\usepackage{algorithm}

\usepackage{array}
\usepackage[caption=false,font=normalsize,labelfont=sf,textfont=sf]{subfig}
\usepackage{textcomp}
\usepackage{stfloats}
\usepackage{url}
\usepackage{verbatim}
\usepackage{xcolor}
\usepackage{graphicx}
\usepackage{xcolor}
\usepackage{newunicodechar}
\usepackage{cases} 
\newunicodechar{，}{,} 
\usepackage{cite}
\usepackage{makecell}
\usepackage{bm}
\begin{document}

\title{Integrated Communication and Control for Energy-Efficient UAV Swarms: A Multi-Agent Reinforcement Learning Approach}

\author{Tianjiao Sun, Ningyan Guo, Haozhe Gu, Yanyan Peng, and Zhiyong Feng,~\IEEEmembership{Senior Member, ~IEEE,}
\thanks{Tianjiao Sun, Ningyan Guo,  Haozhe Gu, Yanyan Peng, and Zhiyong Feng are with the Key Laboratory of Universal Wireless Communications, Ministry of Education, Beijing University of Posts and Telecommunications, Beijing 100876, China (e-mail: suntianjiao@bupt.edu.cn, guoningyan@bupt.edu.cn, Gu$\_$haozhe@bupt.edu.cn, pengyanyan@bupt.edu.cn, fengzy@bupt.edu.cn).}
}



\maketitle

\begin{abstract}
The deployment of unmanned aerial vehicle (UAV) swarm-assisted communication networks has become an increasingly vital approach for remediating coverage limitations in infrastructure-deficient environments, with especially pressing applications in temporary scenarios, such as emergency rescue, military and security operations, and remote area coverage. However, complex geographic environments lead to unpredictable and highly dynamic wireless channel conditions, resulting in frequent interruptions of air-to-ground (A2G) links that severely constrain the reliability and quality of service in UAV swarm-assisted mobile communications. To improve the quality of UAV swarm-assisted communications in complex geographic environments, we propose an integrated communication and control co-design mechanism. Given the stringent energy constraints inherent in UAV swarms, our proposed mechanism is designed to optimize energy efficiency while maintaining an equilibrium between equitable communication rates for mobile ground users (GUs) and UAV energy expenditure. We formulate the joint resource allocation and 3D trajectory control problem as a Markov decision process (MDP), and develop a multi-agent reinforcement learning (MARL) framework to enable real-time coordinated actions across the UAV swarm. To optimize the action policy of UAV swarms, we propose a novel multi-agent hybrid proximal policy optimization with action masking (MAHPPO-AM) algorithm, specifically designed to handle complex hybrid action spaces. The algorithm incorporates action masking to enforce hard constraints in high-dimensional action spaces. Experimental results demonstrate that our approach achieves a fairness index of 0.99 while reducing energy consumption by up to 25$\%$ compared to baseline methods.
\end{abstract}


\section{Introduction}
\IEEEPARstart{U}{nder} the constraints of complex electromagnetic environments, traditional terrestrial wireless communication networks face challenges including degraded propagation channels, constrained topological structures, scarce spectrum resources, and limited coverage \cite{ref1}. In contrast, unmanned aerial vehicle (UAV) swarms effectively overcome the limitations of traditional terrestrial wireless networks by leveraging their three-dimensional mobility \cite{ref2}\cite{ref3}, dynamic networking capability, flexible spectrum reuse, and extended coverage, thereby providing communication services for ground users (GUs) \cite{ref4} -\cite{ref7}. 
Therefore, UAV swarm-assisted communication networks have gained growing popularity in both civilian and military applications, such as emergency rescue, military and security operations, and remote area coverage \cite{ref8}\cite{ref9}.

Prior works have investigated joint communication resource allocation and trajectory control for UAV swarms, primarily using conventional optimization methods such as convex optimization. Qiu et al.\cite{ref10} propose an efficient iterative algorithm for joint optimization of user association and UAV-mounted base station placement, utilizing gradient ascent, dual-domain coordinated descent, and bipartite graph matching techniques. Using a successive convex approximation (SCA) algorithm,  Shen et al.\cite{ref11} address multi-UAV interference coordination through joint trajectory design and power control. Liu et al.\cite{ref12} employ the SCA algorithm to maximize system throughput through joint optimization of vehicle communication scheduling, UAV power allocation, and trajectories. Zhu et al.\cite{ref13} propose a suboptimal solution for joint user clustering, transmit/receive beamforming, and UAV placement. In multi-UAV assisted systems, Zhang et al.\cite{ref14} investigate the co-optimization of UAV trajectories, user association, and beamforming using an alternating optimization algorithm. 

However, these works predominantly use simplified line of sight (LoS) channel models for air-to-ground (A2G) links, which facilitate trajectory optimization and average performance analysis. In reality, complex urban environments \cite{ref15} introduce severe signal obstruction that significantly degrades A2G link quality. When channel states are unknown a priori, deep reinforcement learning (DRL) has demonstrated strong potential for solving dynamic and non-convex UAV communication problems \cite{ref18}\cite{ref19}. Multi-agent reinforcement learning (MARL) is particularly suitable for UAV swarm-assisted communication and control systems. Existing MARL methods are broadly categorized by action space nature: discrete or continuous.   

\begin{itemize}\item[$\bullet$]Discrete Action Spaces: Zhong et al. \cite{ref20} propose a mutual deep Q-network (MDQN) algorithm for joint 3D trajectory and power allocation optimization in UAVs. Won Joon Yun et al. \cite{ref21} investigate a model-free MARL-based collaborative scheme for autonomous surveillance UAVs, targeting energy-efficient and reliable surveillance. Wang et al. \cite{ref22} develop a Q-value mixing (QMIX)-based algorithm to jointly optimize the trajectories of UAVs and sensor node scheduling, minimizing the time-average total expected age of information.

\item[$\bullet$]Continuous Action Spaces: In a practical user roaming scenario served by multiple UAVs \cite{ref23}, a hybrid reward multi-agent proximal policy optimization (HR-MAPPO) algorithm is proposed to jointly optimize UAVs’ trajectory and beamforming to maximize the sum data rate. Xu et al. \cite{ref24} investigate  multi-UAV trajectory design for uplink data collection task using the Q-network in DRL. In  \cite{ref25} and \cite{ref26}, multiple UAVs are deployed to serve ground users on continuous maps. The former introduces the multi-agent long short-term memory-deep deterministic policy gradient (LSTM-DDPG) method to maximize data collection, while the latter utilizes the proximal policy optimization (PPO) to minimize the cumulative content acquisition delay for users. 
\end{itemize}

Nevertheless, the aforementioned studies predominantly address discrete or continuous action spaces in isolation, often overlooking the complexities inherent to hybrid action space scenarios. This limitation substantially constrains the decision-making capacity of UAV swarms in dynamic real-world environments. Although the latest work \cite{ref27} investigates the UAV data collection with discrete variables (radar, communication, movement) and continuous variables (transmit power, flying direction, velocity), it ignores the obstruction impact on A2G links in complex geographic environment, which is a critical factor for reliable communication performance. 


Existing research on UAV swarm-assisted integrated communication and control systems still faces critical technical challenges in practical deployment. First, the unpredictability and dynamics of the channel in real-world lead to frequent interruptions of the A2G link, which seriously degrades communication fairness and quality of service. Second, UAV swarms are limited by current decision-making capabilities and lack essential mechanisms to coordinate their actions. To address these, a novel MARL framework is proposed to enable autonomous decision-making in UAV swarm-assisted integrated communication and control systems. The main contributions of this paper are summarized as follows:

\begin{enumerate} \item[1)] To enhance quality of communication service in complex geographic environment, we propose a UAV swarm-assisted integrated communication and control co-design mechanism. Considering the energy constraints of UAV swarms, we formulate an objective function that characterizes the trade-off between the fairness-constrained communication rates for ground users and UAVs’ energy consumption. \item[2)] We model the UAV swarm’s resource allocation and 3D trajectory control as a Markov decision process (MDP) and propose a MARL framework as a solution, which facilitates decision-making on collaborative actions among UAVs in real time. \item[3)] To address hybrid action space challenges in UAV swarms, we propose a  novel multi-agent hybrid proximal policy optimization with action masking (MAHPPO-AM) algorithm that guarantees hard constraints in high-dimensional action spaces via action masking. \item[4)] Experimental results demonstrate that our approach achieves a fairness index of 0.99 while reducing energy consumption by up to 25$\%$ compared to baseline methods.  \end{enumerate}

The remainder of this paper is organized as follows. Section II presents the system model and problem formulation. In Section III, Markov decision process (MDP) is formally defined and we present the MARL-based solution to the optimization problem. Simulation results are presented and discussed in Section IV. In Section V, we conclude the whole paper.

\section{System Model and Problem Formulation}
\subsection{System Model}
\begin{figure}[!t]
\centering
\includegraphics[width=3.5in]{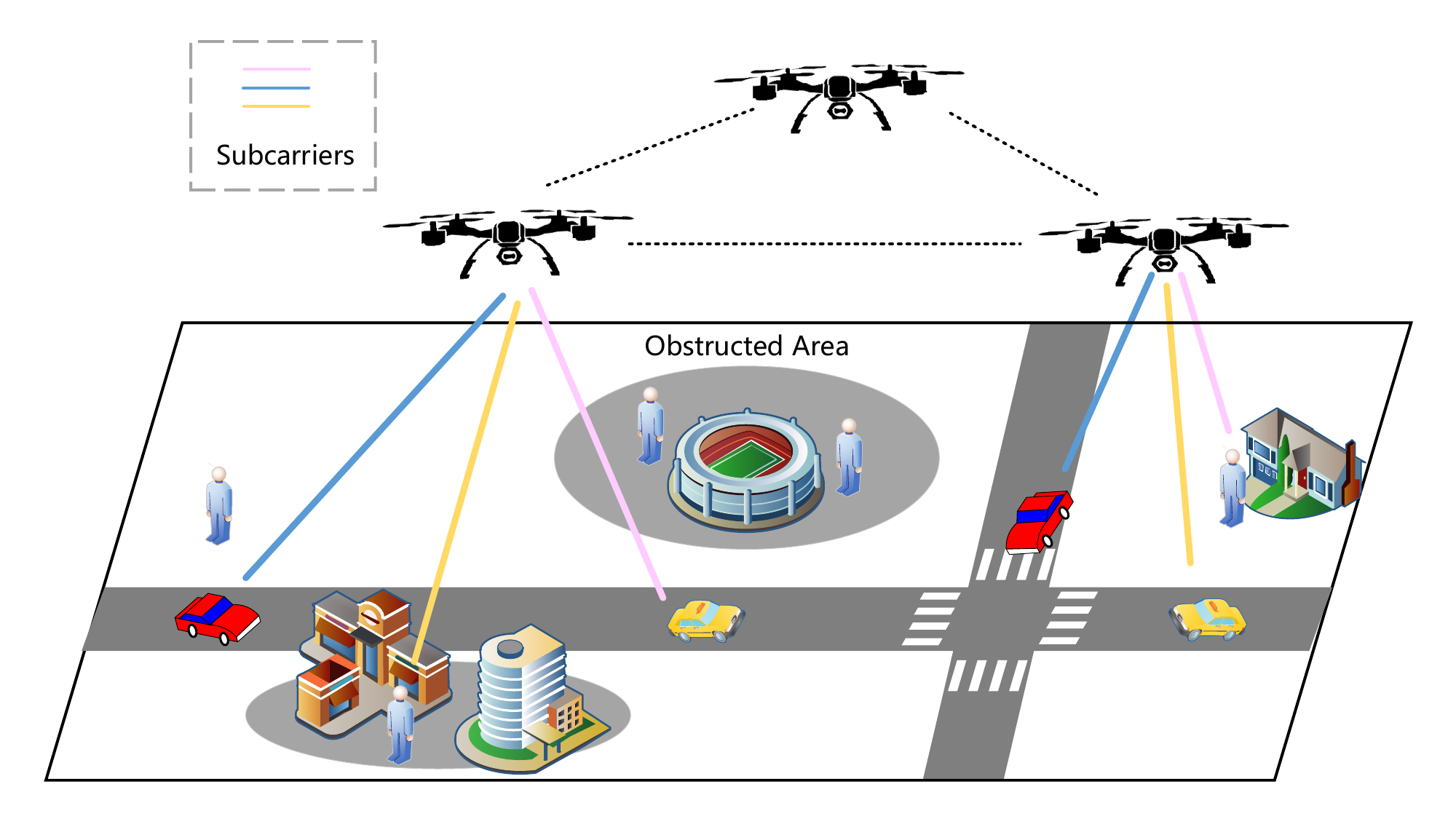}
\caption{UAV swarm-assisted communication network.}
\label{fig1}
\end{figure}
As shown in Fig.\ref{fig1}, we focus on a UAV swarm-assisted orthogonal frequency division multiple access (OFDMA) downlink communication network in obstructed outdoor environments, where $M$ UAVs are deployed as aerial base stations to serve $K$ mobile ground users via $N$ orthogonal subcarriers for each UAV. The sets of UAVs, users, and subcarriers are denoted as ${\cal M} = \{ 1,2,...,M\} $, ${\cal K} = \{ 1,2,...,K\} $, and ${\cal N} = \{ 1,2,...,N\} $. It is assumed that channel resources are insufficient to serve all the users simultaneously, i.e., $MN \le K$, making resource allocation critical for enhancing system performance. Each UAV allocates orthogonal subcarriers to its served ground users, eliminating intra-UAV interference. However, since all the UAVs share the same $N$ subcarriers, users served by different UAVs on the same subcarrier may suffer from mutual interference.  
 
Meanwhile, buildings and other obstacles may cause severe obstruction to the A2G links and drastically weaken the received signal strength of users in obstructed area. To ensure that UAV swarms can effectively serve these users, it is essential to design UAV trajectory that maximize downlink communication performance. We consider the UAVs’ trajectory and resource allocation by discretizing the entire time horizon into $T$ equally-spaced time slots. The duration of each timeslot is ${T_s}$. In a 3D Cartesian coordinate, the instantaneous location of UAV $m$ within timeslot $t$ is denoted by ${q_m}(t) = {[{x_m}(t),{y_m}(t),{h_m}(t)]^T}$, where $\bm{q}(t) = \{ {q_1}(t),...,{q_M}(t)\} $ with $t \in {\cal T} = \{ 1,...,T\} $. The horizontal coordinate of ground user $k$ is denoted by ${u_k}(t) = {[{x_k}(t),{y_k}(t)]^T}$.

\textit{1) UAV-GU Communication Model}
\ 
\newline 
\indent According to the 3GPP Release 15 \cite{ref28}, we can obtain the A2G channel model between each UAV and ground user. The path loss depends on LoS and non-line-of-sight (NLoS) link states and the design formulas of path loss 
${L_{LoS/NLoS}}(t)$ between the user $k$ and UAV $m$ can be expressed as (\ref{eq1}), shown at the bottom of the page, where ${f_c}$ denotes carrier frequency, and the 3D distance from UAV $m$ to user $k$ at timeslot $t$ is denoted as (\ref{eq2}).
\begin{figure*}[b]
    \centering

    \vspace{0.5em}
    \hrule width 1\linewidth height 0.4pt
    \vspace{0.5em}
    \begin{equation}
        {L_{LoS/NLoS}}(t) = \left\{ \begin{array}{l}
        30.9 + (22.25 - 0.5{\log _{10}}{h_m}(t)){\log _{10}}d_k^m(t) + 20{\log _{10}}{f_c},\;\;\;\;\;\;\;\;\;\;\;\;\;\;\;\;\;\;\;{\rm{if}}\;\;LoS,\\
        \max \{ {L_{LoS}},32.4 + (43.2 - 7.6{\log _{10}}{h_m}(t)){\log _{10}}d_k^m(t) + 20{\log _{10}}{f_c}\} ,\;\;{\rm{if}}\;\;NLoS.\;
        \end{array} \right.
        \label{eq1}
    \end{equation}
    
    \begin{equation}
        d_k^m(t) = \sqrt {{h_m}^2(t) + {{[{x_m}(t) - {x_k}(t)]}^2} + {{[{y_m}(t) - {y_k}(t)]}^2}} .
        \label{eq2}
    \end{equation}
    
    \begin{equation}
        {P_{LoS}}(t) = \left\{ \begin{array}{l} 
        1,\;\;\;\;\;\;\;\;\;\;\;\;\;\;\;\;\;\;\;\;\;\;\;\;\;\;\;\;\;\;\;\;\;\;\;\;\;\;\;\;\;\;\;\;\;\;\;\;\;\;\;\;\;\;\;\;\;\;\;\;\;\;\;\;\;\;\;\;\;\;\;\;\;\;\;\;\;\;\;\;\;\;\;\;\;{\rm{if}}\;\sqrt {{{(d_k^m(t))}^2} - {{(h_m^{}(t))}^2}}  \le {d_0},\\
        \frac{{{d_0}}}{{\sqrt {{{(d_k^m(t))}^2} - {{(h_m^{}(t))}^2}} }} + \exp \left\{ {\left. {\frac{{ - \sqrt {{{(d_k^m(t))}^2} - {{(h_m^{}(t))}^2}} }}{{{p_1}}} + \frac{{{d_0}}}{{p{}_1}}} \right\}} \right.,\;\;\;{\rm{if}}\sqrt {{{(d_k^m(t))}^2} - {{(h_m^{}(t))}^2}}  > {d_0}.
        \end{array} \right.
        \label{eq3}
    \end{equation}
\end{figure*}

In the propagation model, the probability of LoS is denoted as ${P_{LoS}}(t)$ in (\ref{eq3}), shown at the bottom of the page, where ${d_0} = \max [294.05 \cdot {\log _{10}}{h_m}(t) - 432.94,\;18]$ and  ${p_1} = 233.98 \cdot {\log _{10}}{h_m}(t) - 0.95$. Logically, the NLoS probability is ${P_{NLoS}}(t) = 1 - {P_{LoS}}(t)$. Thus, the mean path loss between the UAV $m$ and user $k$ can be given by
    \begin{equation} 
        L_k^m(t) = {P_{LoS}} \cdot {L_{LoS}} + {P_{NLoS}}{L_{NLoS}}.
        \label{eq4}
    \end{equation}
Considering small-scale fading, the channel gain from the UAV $m$ and user $k$ at timeslot $t$ can be calculated by  
    \begin{equation}
        g_k^m(t) = H_k^m(t) \cdot {10^{ - L_k^m(t)/10}},
        \label{eq5} 
    \end{equation}
where  $H_k^m(t)$ is the fading coefficient \cite{TWC-2018} between UAV $m$ and user $k$.

To facilitate formulation, we define a set of binary variables $\bm{\phi}(t) = \{ {\phi _{k,m,n}}(t)|k \in {\cal K},m \in {\cal M},n \in {\cal N}\} $ as the indicators for resource allocation involving channel assignment and the user association, where ${\phi _{k,m,n}}(t) = 1$ refers to user $k$ served by UAV $m$ on channel $n$ at timeslot $t$ and 
${\phi _{k,m,n}}(t) = 0$ means otherwise. Note that $\bm{p}(t) = \{ {p_{m,n}}(t)|m \in {\cal M},n \in {\cal N}\}$ is the set of transmit power of the UAV swarm, and ${p_{m,n}}(t)$ denotes the downlink transmit power of UAV $m$ on channel $n$ within timeslot $t$. Hence, in timeslot $t$, signal-to-interference-plus-noise ratio (SINR) received by user $k$ from UAV $m$ on channel $n$ can be expressed by 
    \begin{equation}
        {\gamma _{k,m,n}}(t) = \frac{{{p_{m,n}}(t){{\left| {g_k^m(t)} \right|}^2}}}{{\sum\limits_{j = 1,j \ne m}^M {{p_{j,n}}(t){{\left| {g_k^j(t)} \right|}^2} + {\sigma ^2}} }},
        \label{eq6} 
    \end{equation}
where ${\sigma ^2}$ denotes the noise power and the term ${\sum\nolimits_{j = 1,j \ne m}^M {{p_{j,n}}(t)\left| {g_k^j(t)} \right|} ^2}$ in the denominator refers to the co-channel interference. Therefore, the link rate achieved by user $k$ at timeslot $t$ \cite{ref15} is given by
    \begin{equation}
        {R_k}(t) = \sum\limits_{m = 1}^M {\sum\limits_{n = 1}^N {{\phi _{k,m,n}}(t)\log \left( {1 + \gamma _{k,m,n}(t)} \right)} } .
        \label{eq7} 
    \end{equation}

The total achievable rate by all the ground users at timeslot $t$ is 
    \begin{equation}
        {R_c}(t) = \sum\limits_{k = 1}^K {{R_k}(t)} .
        \label{eq8}  
    \end{equation} 
Thus, the total achievable rate before timeslot $t$ is
    \begin{equation} 
        \bar R(t) = {R_c}(1) + ... + {R_c}(t),
        \label{eq9}  
    \end{equation} 
and the achievable rate by the ground user $k$ before timeslot $t$ is
    \begin{equation} 
        {\bar R_k}(t) = {R_k}(1) + ... + {R_k}(t).
        \label{eq10}  
    \end{equation} 

However, maximizing the total transmission rate may lead to inequity, where the UAV may tend to be close to some users with good channel conditions, while obstructed users suffer from low rate all the time. To tackle this issue, we refer to \cite{ref29} and define the rate ratio of user $k$ as
    \begin{equation} 
        {f_k}(t) = \frac{{{{\bar R}_k}(t)}}{{\bar R(t)}}.
        \label{eq11}  
    \end{equation} 
Then Jain’s fairness index \cite{ref30} is employed to measure the fairness among users
    \begin{equation} 
        \hat f(t) = \frac{{{{\left( {\sum\nolimits_{k = 1}^K {{f_k}(t)} } \right)}^2}}}{{K\left( {\sum\nolimits_{k = 1}^K {{f_k}{{(t)}^2}} } \right)}}.
        \label{eq12}  
    \end{equation} 
where $\hat f(t) \in [0,1]$. The smaller the differences among the rate ratios ${\{ {f_k}(t)\} _{k \in {\cal K}}}$ are, the greater fairness index $\hat f(t)$ is. Notably, both the fairness index and the total rate are the functions of the UAV swarm trajectory $\bm{q}$, transmit power $\bm{p}$, and channel allocation $\bm{\phi}$. Then the total fair rate during the whole mission can be defined as
    \begin{equation} 
        R(t) = \hat f(t){R_c}(t).
        \label{eq13}  
    \end{equation} 

\textit{2) UAV Energy Consumption Model}
\ 
\newline 
\indent  UAV swarm-assisted communication systems face the challenge of energy limitations, and the energy consumption of the UAV consists of two parts: one for communication and the other for propulsion energy to generate thrusts. In practice, the communication energy is often much smaller than flight energy by two orders of magnitude \cite{ref31}. As a result, the propulsion energy consumption of the UAV must be taken into account. 

For clarity, we consider only the acceleration component aligned with the UAV’s velocity and ignore the component perpendicular to it. Under this assumption, it is reasonable to allow the UAV to change its flight direction instantaneously without incurring additional energy consumption, as a quadrotor UAV can easily maneuver by adjusting the rotational speed of its four rotors. The thrust of rotors is a function of UAV velocity ${\bm{v}_m}(t)$ and acceleration $\bm{a}_m^c(t)$. And ${v_m}(t) = \left\| {\bm{v}_m}(t) \right\|$ is the UAV speed. Then the thrust of UAV $m$ can be expressed as 
    \begin{equation}  
        ||F(\bm{v}_m)|| = \left\| {\left. {m'\bm{a}_m^c + \frac{1}{2}\rho {S_{FP}}||\bm{v}_m||\bm{v}_m - m'\bm{g}} \right\|} \right.,   
        \label{eq14}  
    \end{equation} 
where $m'$ denotes the UAV mass. $\rho $ and ${S_{FP}}$ are the air density and fuselage equivalent flat plate area, respectively. $\bm{g}$ denotes the gravity acceleration vector. Considering ${\bm{v}_m}(t)\buildrel \Delta \over = {\dot q_m}(t)$ and $\bm{a}_m^c(t) \buildrel \Delta \over = {\ddot q_m}(t)$, ${\dot q_m}(t)$ and ${\ddot q_m}(t)$ denote the first- and second-order derivative with respect to $t$, respectively. The propulsion power is essentially a function of the UAV trajectory ${q_m}(t)$. Then with a given trajectory ${q_m}(t)$, the propulsion energy of UAV $m$ at a timeslot can be expressed as (\ref{eq15}), shown at the bottom of the page, 
\begin{figure*}[b]
    \centering

    \vspace{0.5em}
    \hrule width 1\linewidth height 0.4pt
    \vspace{0.5em}
    \begin{equation}
        \begin{array}{l}
        {E_m}(t) = \left( {\frac{\delta }{8}\left( {\frac{{||F({{\dot q}_m},{{\ddot q}_m})||}}{{{c_T}\rho A}} + 3||\bm{v}_m|{|^2}} \right)\sqrt {\frac{{\rho c_S^2A||F({{\dot q}_m},{{\ddot q}_m})||}}{{{c_T}}}}  + m'||\bm{g}||||\bm{v}_m||\sin {\tau _c}} \right){T_s} + \\
        \;\;\;\;\;\;\;\;\;\;\;\left( {(1 + {c_f})||F({{\dot q}_m},{{\ddot q}_m})||{{\left( {\sqrt {\frac{{||F({{\dot q}_m},{{\ddot q}_m})|{|^2}}}{{4{\rho ^2}{A^2}}} + \frac{{||\bm{v}_m|{|^4}}}{4}}  - \frac{{||\bm{v}_m|{|^2}}}{2}} \right)}^{\frac{1}{2}}} + \frac{1}{2}\rho {S_{FP}}||\bm{v}_m|{|^3}} \right){T_s}.
        \end{array}
        \label{eq15}
    \end{equation}
\end{figure*}
where $\delta $ denotes the local blade section drag coefficient; ${c_T}$ is the thrust coefficient based on disc area; $A$ and ${c_S}$ denote the disc area and rotor solidity, respectively; ${c_f}$ is the incremental correction factor of induced power; ${\tau _c}$ denotes the climb angle. Further, the total propulsion energy consumption of the UAV swarm is given by
    \begin{equation} 
        E(t) = \sum\limits_{m = 1}^M {{E_m}(t)} .
        \label{eq16}  
    \end{equation} 

\subsection{Problem Formulation}
In this paper, we propose a resource allocation and trajectory control mechanism for UAV swarms in 3D space. Specifically, the total achievable rate ${R_c}(t)$ of mobile ground users is first modeled as a function of the subcarrier-user correlation coefficient ${\phi _{k,m,n}}(t)$, the UAV subcarrier transmit power ${p_{m,n}}(t)$, and the UAV trajectory ${q_m}(t)$ based on the A2G channel model specified by the 3GPP protocol. Subsequently, the UAV energy consumption ${E_m}(t)$ is modeled as a function of UAV trajectory ${q_m}(t)$. Due to the size and weight constraints, on-board batteries with limited energy will lead to endurance and performance degradation. Thus, energy efficiency, defined as the information bits per unit energy consumption, is a critical issue in UAV swarm-assisted communication and control. Finally, the optimization objective of this paper is to maximize the long-term energy efficiency of the UAV swarm, which can be formulated as 
\begin{IEEEeqnarray}{lll}
    \IEEEyesnumber\label{eqP1}
    \text{(P1):} & \mathop{\max}\limits_{\bm{\phi},\bm{p},\bm{q}} \sum_{t=1}^T \eta^{ee}(t) \\
    \text{s.t.} & ||q_m(t) - q_j(t)|| \geq d_{\min}, \quad m,j \in \mathcal{M},\ m \neq j, & \label{eqa} \\
    & 0 \leq v_m(t) \leq v_{\max}, &  \label{eqb} \\
    & 0 \leq ||\bm{a}_m^c(t)|| \leq a_{\max}, & \label{eqc} \\
    & h_{\min} \leq h_m(t) \leq h_{\max}, & \label{eqd} \\
    & \sum_{k=1}^K \phi_{k,m,n}(t) \leq 1, &  \label{eqe} \\
    & \sum_{m=1}^M \sum_{n=1}^N \phi_{k,m,n}(t) \leq 1, & \label{eqf} \\
    & \sum_{n=1}^N p_{m,n}(t) \leq p_{\max}, & \label{eqg}
\end{IEEEeqnarray}
where    
    \begin{equation} 
        {\eta ^{ee}}(t) = \frac{{\hat f(t){R_c}(t)}}{{E(t)}}.
        \label{eqp1}  
    \end{equation} 
The optimization objective captures the trade-off between ensuring fair communication rates for mobile users and minimizing the UAVs’ energy consumption. In practice, constraint (\ref{eqa}) enforces a safe distance between UAVs to prevent collisions. Constraints (\ref{eqb}), (\ref{eqc}), and (\ref{eqd}) indicate the limitations of UAVs’ flying speed, acceleration, and height. Constraints (\ref{eqe}) and (\ref{eqf}) ensure that each UAV serves at most one user on one subcarrier and each user can be only served by at most one UAV. Constraint (\ref{eqg}) limits the total power budget of each UAV. Apparently, the optimization objective is mixed-integer and non-convex, which is generally NP-hard. As such, conventional convex optimization methods are inadequate for solving it in real time. To address this challenge, we propose a multi-agent reinforcement learning framework as an effective solution.

\section{Multi-Agent Reinforcement Learning-based Solutions}
In this section, the UAV swarm’s resource allocation and 3D trajectory control are considered as a sequential decision process, i.e., a decision is made at each single step based on the current observation. Under the MARL framework, UAVs play the role of agents whose goal is to maximize long-term energy efficiency while ensuring fair communication.

\subsection{Markov Decision Process Definitions}
Mathematically, the complete MDP can be denoted by a tuple $\{ {\cal S},{\cal A},{\cal P},r,\gamma \} $, where ${\cal S}$ is the state space, ${\cal A}$ is the action space, ${\cal P}$ is the state transition probability, $r$ is the reward at each step, and $\gamma $  is the discount factor. In each timeslot, the UAV $m$ chooses an action ${a_m}$ with policy ${\pi _m}({a_m}|{s_m})$. The joint action $a \in {\cal A}$ of all the UAVs causes an environment transition from the current state $s \in {\cal S}$ to the next state $s' \in {\cal S}$ following the state transition  $P(s'|s,a)$, which is unknown to all the UAVs. And the mutual goal of all the UAVs is to maximize the cumulative discounted total return
    \begin{equation} 
        G = \sum\limits_{t = 1}^T {{\gamma ^{t - 1}}{r_t}} .
        \label{eq26}  
    \end{equation} 

As discussed in Section II-B, (P1) involves combinatorial optimization and is highly coupled between the variables, which makes it difficult for traditional convex optimization to solve in real-time. Fortunately, multi-agent RL algorithms allow each agent to make an optimal decision following its learned policy through a trial-and-error way. To fit the problem of resource allocation and trajectory control into a MARL framework, the essential elements of the Markov decision process can be defined as follows.   

\textit{1) State Space}
\ 
\newline 
\indent  As mentioned in Section II-A, the link rate is related to the path loss between the UAV and mobile GUs. However, the locations of the UAVs and GUs are easier to obtain than path loss, as most mobile GUs are equipped with global positioning system (GPS) sensors. Besides, the locations of GUs change all the time. As a result, the state space includes the GUs’ locations $\{ {u_k}(t),k \in {\cal K}\} $ and the UAVs’ locations $\{ {q_m}(t - 1),m \in {\cal M}\}$ from the previous timeslot such that UAVs can deal with the movement of GUs. Moreover, since the flight energy consumption of a UAV is a function of the UAV’s speed, the state includes the UAVs’ speed $\{ v_m^{}(t - 1),m \in {\cal M}\}$ from the previous timeslot, which also aims to remind the UAV not to exceed the acceleration constraint. Thus, the state can be defined as 
    \begin{equation} 
        s(t) = \{ {u_k}(t),{q_m}(t - 1),v_m^{}(t - 1)|k \in {\cal K},m \in {\cal M}\} ,
        \label{eq27}  
    \end{equation} 
and has $3(K + M + 1)$ dimensions. 

\textit{2) Hybrid Action Space}
\ 
\newline 
\indent  The action decided by the UAV $m$ consists of user-channel assignment，transmit power allocation, and UAV’s trajectory, which can be defined as  
    \begin{equation} 
        {a_m}(t) = \{ {\phi _{k,m,n}}(t),{p_{m,n}}(t),{q_m}(t)|k \in {\cal K},m \in {\cal M},n \in {\cal N}\} .
        \label{eq28}  
    \end{equation} 
\begin{itemize}\item[$\bullet$]User-Channel Assignment: ${{\phi _{k,m,n}}(t)} \in \{ 0,1\}$ denotes a binary discrete variable，and  ${\phi _{k,m,n}}(t) = 1$ refers to the user $k$ served by UAV $m$ on channel $n$ at timeslot $t$. The user-channel assignment for UAV $m$ can be expressed as (\ref{eq29}), shown at the bottom of the next page, where $\sum\limits_{k = 1}^K {{\phi _{k,m,n}}(t)}  \le 1$ and $\sum\limits_{m = 1}^M {\sum\limits_{n = 1}^N {{\phi _{k,m,n}}(t)}  \le 1}$.
\begin{figure*}[b]
    \centering

    \vspace{0.5em}
    \hrule width 1\linewidth height 0.4pt
    \vspace{0.5em}
    \begin{equation} 
        {\phi _m} = \left[ \begin{array}{l}
        {\phi _{1,m,1}}\;\;\;\;{\phi _{1,m,2\;}}\;\;\;...\;\;\;{\phi _{1,m,N - 1\;}}\;\;\;\;{\phi _{1,m,N\;}}\\
        \;\;...\;\;\;\;\;\;\;\;...\;\;\;\;\;\;...\;\;\;\;\;\;\;...\;\;\;\;\;\;\;\;\;...\\
        {\phi _{K,m,1}}\;\;\;{\phi _{K,m,2\;}}\;\;...\;\;\;{\phi _{K,m,N - 1\;}}\;\;\;{\phi _{K,m,N\;}}
        \end{array} \right],
        \label{eq29}  
    \end{equation} 

    \begin{equation} 
        {a_m}(t) = \{ {\phi _{k,m,n}}(t),{p_{m,n}}(t),{v_m}(t),{\theta _m}(t),{\varphi _m}(t)|k \in {\cal K},m \in {\cal M},n \in {\cal N}\} .
        \label{eq30}  
    \end{equation} 
\end{figure*}

\item[$\bullet$]Transmit Power Allocation: $\{ {p_{m,n}}(t)|m \in {\cal M},n \in {\cal N}\}$ denotes the downlink transmit power of UAV $m$ on channel $n$ within timeslot $t$ and ${p_{\max }}$ denotes the UAV’s total power budget in each timeslot. We also assume that the total power budget of each UAV is limited, i.e., ${\rm{ }}\sum\limits_{n = 1}^N {{p_{m,n}}(t)}  \le {p_{\max }}$. 
\item[$\bullet$]UAV’s Trajectory: In 3D space, we utilize the spherical coordinate $({v_m}(t),\;\theta (t),\;\varphi (t))$ to describe the UAV’s speed and flight direction more conveniently, where ${v_m}(t) \in (0,\;{v_{\max }})$ is the UAV’s speed, $\theta (t)  \in (0,2\pi )$ is the azimuthal angle, and $\varphi(t)  \in (\pi /3,\;2\pi /3)$ is known to be the polar angle. Such a definition always satisfies the constraints of the UAV’s maximum speed and pitch angle in each time slot.  
\end{itemize}

In summary, the action of UAV $m$ actually consists of five parts (\ref{eq30}), shown at the bottom of the next page. Thus, the joint action space of all UAVs can be expressed as,  
    \begin{equation} 
        a(t) = \{ {a_1}(t),...,{a_M}(t)\} .
        \label{eq31}  
    \end{equation} 

\textit{3) Reward}
\ 
\newline 
\indent  In MARL, reward serves as a measure of how favorable an action is under a given state. By appropriately designing the reward function, the original non-convex objective (P1) can be transformed into a problem of maximizing cumulative reward. Given that we aim to maximize long-term energy efficiency while ensuring fair communication, the immediate reward can be defined as follows,
    \begin{equation} 
        r(t) = {\eta ^{ee}}(t).
        \label{eq32}  
    \end{equation} 

\subsection{Design of the MAHPPO-AM Algorithm}
In this section, we propose an MAHPPO-AM algorithm to solve the UAV swarm-assisted integrated communication and control co-design. We first design the action masking of the algorithm and then provide the optimization process of the multi-agent RL algorithm. 

\textit{1) Action Masking}
\ 
\newline  
\indent  In MARL, the purpose of action masking is to restrict the agent's action space to exclusively select valid actions that conform to the rules and constraints of the environment \cite{ref32}. Hence, the approach is especially critical for discretized user-channel assignment actions. On the one hand, the mechanism constructs a dynamic feasible action space by means of predefined constraint rules. Compared with conventional policy networks that directly output unfiltered action distributions, this constraint-embedded approach fundamentally prevents invalid allocation, such as multiple users on the same channel or overcapacity resource allocation. Consequently, this approach ensures that the output decisions meet communication fairness requirements at each timeslot.  

On the other hand, the exploration space of agent is compressed to the valid policy subspace by masking invalid actions. This mechanism not only reduces the sample waste of invalid exploration, but also accelerates the gradient optimization process by reducing the policy search dimension. Based on real-time environmental observations, action masking is designed to satisfy hard constraints in high-dimensional discrete action spaces. Specifically, it ensures that user-channel assignment $\bm{\phi} $ satisfies the following constraints:
\begin{itemize}\item[$\bullet$]Global GU Constraint: Each user is served by at most one UAV on a single channel, i.e., $\sum\limits_{m = 1}^M {\sum\limits_{n = 1}^N {{\phi _{k,m,n}}(t)}  \le 1} .$
\item[$\bullet$]Global Channel Constraint: Each channel is assigned to a maximum of $M$ UAVs, i.e., $\sum\limits_{m = 1}^M {\sum\limits_{k = 1}^K {{\phi _{k,m,n}}(t)}  \le M}$.
\item[$\bullet$]UAV Constraint: Each UAV serves at most one user or on one subcarrier，i.e., $\sum\limits_{k = 1}^K {{\phi _{k,m,n}}(t)}  \le 1,\sum\limits_{n = 1}^N {{\phi _{k,m,n}}(t)}  \le 1.$
\end{itemize}

The dynamic valid action space is defined as ${\tilde {\cal V}_t} \subseteq {\cal K} \times {\cal M} \times {\cal N}$, and elements $(k,m,n)$ satisfy (\ref{eq33}), shown at the bottom of the page.
\begin{figure*}[b]
    \centering

    \vspace{0.5em}
    \hrule width 1\linewidth height 0.4pt
    \vspace{0.5em}
    \begin{equation} 
        {\tilde {\cal V}_t} = \left\{ {\left. {(k,m,n)|\underbrace {\sum\limits_{m' = 1}^M {\sum\limits_{n' = 1}^N {{\phi _{k,m',n'}}(t) = 0} } }_{{\rm{Global}}\;{\rm{GU}}\;{\rm{Constraint}}} \wedge \underbrace {\sum\limits_{m' = 1}^M {\sum\limits_{k' = 1}^K {{\phi _{k',m',n}}(t) < M} } }_{{\rm{Global}}\;{\rm{Channel}}\;{\rm{Constraint}}} \wedge \underbrace {\sum\limits_{n' = 1}^N {{\phi _{k,m,n'}}(t)}  = 0}_{{\rm{UAV - GU}}\;{\rm{Constraint}}} \wedge \underbrace {\sum\limits_{k' = 1}^K {{\phi _{k',m,n}}(t)}  = 0}_{{\rm{UAV - Channel}}\;{\rm{Constraint}}}} \right\}} \right..
        \label{eq33}  
    \end{equation} 
\end{figure*}
Iterating over $(k,m,n) \in {\cal K} \times {\cal M} \times {\cal N}$, those satisfying ${\tilde {\cal V}_t}$ are filtered to form a candidate action set ${{\cal V}_t}$. For each $\left( {k,m,n} \right) \in {{\cal V}_t}$, the normalized probability can be calculated by
    \begin{equation} 
        P(k,m,n) = \frac{{{e^{{z_{k,m,n}}}}}}{{\sum\nolimits_{(k',m',n') \in {{\cal V}_t}} {{e^{{z_{k',m',n'}}}}} }},
        \label{eq34}  
    \end{equation} 
 where ${z_{k,m,n}}$ is the output logits of the $m$-th agent’ policy network. The action masking restricts the policy search to the valid space ${\tilde {\cal V}_t} \subseteq {\cal K} \times {\cal M} \times {\cal N}$ by dynamically filtering invalid actions, thus supporting the constrained coupling of multi-agent collaborative decision-making.

\begin{figure*}[t] 
\centering
\includegraphics[width=\textwidth]{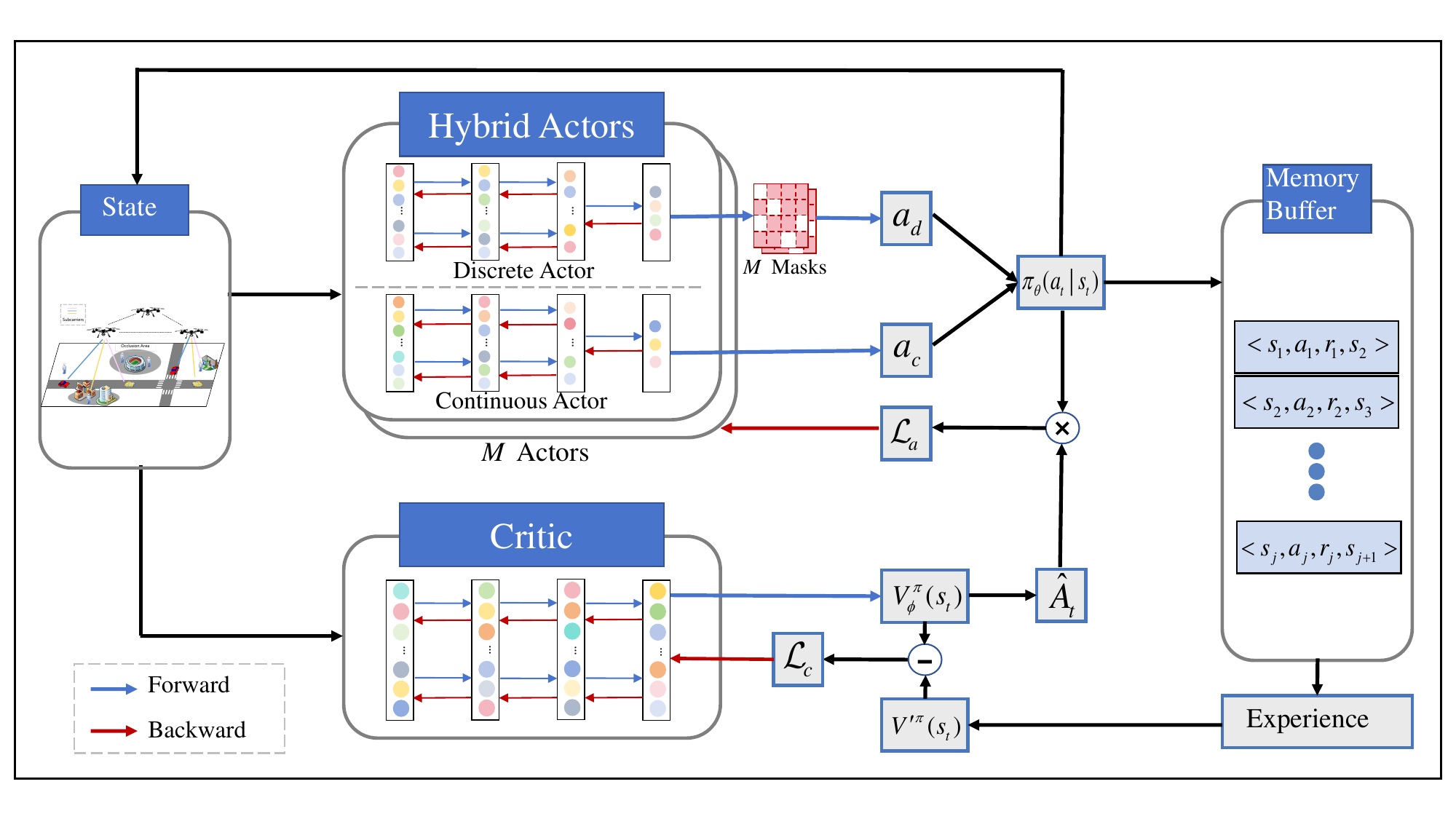}
\caption{The framework of MAHPPO-AM algorithm. The algorithm consists of multiple actor networks and a global critic network. All state vectors observed from the environment are concatenated and sent to the actors and critic, respectively. Each actor outputs discrete actions ${a_d}$ and continuous actions ${a_c}$ for its corresponding agent. The critic outputs the predicted state value. The memory buffer stores previous experience. }
\label{fig2} 
\end{figure*}

\textit{2) MAHPPO-AM Algorithm}
\ 
\newline 
\indent  Traditional PPO algorithms are difficult to deal with both discrete and continuous action spaces. In order to solve this problem, we investigate a multi-agent hybrid proximal policy optimization algorithm to address the UAV swarm-assisted integrated communication and control co-design. As shown in Fig. \ref{fig2}, the proposed MAHPPO-AM algorithm adopts an Actor-Critic structure. The UAV swarm-assisted scenario contains multiple agents, so multiple actor networks are used to provide behavioral decisions for the agents, and the number of actor networks is equal to the number of $M$ agents. To effectively tackle the hybrid action space, we add output branches for the corresponding continuous and discrete actions in each actor network to obtain discrete actions ${a_d}$ and continuous actions ${a_c}$. The five branches share several front layers that encode state information and output user-channel assignment, transmit power, UAV’s speed, azimuth, and polar angle, respectively. The actor branches that provide discrete actions output the probability of choosing different possible actions simultaneously, which is achieved by adding a softmax function to the end of these branches. At the same time, the actor branches that provide continuous actions output the mean and standard deviation of the actions, and the continuous actions are Gaussian distributed.   

In the MAHPPO-AM algorithm, $\phi$ and ${\theta _m}$ are denoted as the parameters of the critic network and the actor network, where ${\theta _m}$ denotes the parameter of the actor network corresponding to the $m$-th agent. In the UAV swarm-assisted integrated communication and control system, the inputs to the Actor-Critic network consist of state vectors, which are GUs’ locations, UAVs’ locations, and UAVs’ speed. In practice, these state vectors are concatenated into a single vector. The actor network outputs the policy $\pi _\theta ^{}({a_t}|{s_t})$, which is the distribution of the action ${a_t}$ predicted at the state ${s_t}$. The critic network outputs the predicted state value $V_\phi ^\pi ({s_t})$, which is the expected cumulative return of state ${s_t}$ and guides the update of all actor networks during the training phase.    

First, we introduce the objective of the critic network in the algorithm. The experience is obtained through a Markov decision process, which describes the interaction between the agents and the environment in the multi-agent RL. We can then obtain the reward for each timeslot in the experience, and the real cumulative return at state ${s_t}$ is given by
    \begin{equation} 
        {V'}^\pi ({s_t}) = \sum\limits_{t' = t}^T {{\gamma ^{t' - t}}{r_{t'}}} ,
        \label{eq35}  
    \end{equation} 
where $\gamma  \in [0,1]$ denotes a discount factor that balances long-term and short-term return. The sampled value $V_\phi ^\pi ({s_t})$ is used as the expected value of the cumulative return to train the critic network, which has the following loss function
    \begin{equation} 
        {{\cal L}_c}(\phi ) = ||V_\phi ^\pi ({s_t}) - {V'}^\pi ({s_t})|{|_2}.
        \label{eq36}  
    \end{equation} 

Then, we present the objective of the actor network. According to the trust region policy optimization \cite{ref33}, we maximize the objective as follows
    \begin{equation} 
        \mathop {\max }\limits_\theta  {\mathbb{E}_t}\left[ {\frac{{\pi _\theta ^{}({a_t}|{s_t})}}{{\pi _{\theta old}^{}({a_t}|{s_t})}}{{\hat A}_t}} \right],
        \label{eq37}  
    \end{equation} 
where $\pi _\theta ^{}({a_t}|{s_t})$ is the current policy and $\pi _{{\theta _{old}}}^{}({a_t}|{s_t})$ is the old policy for collecting trajectories. In addition, the advantage function ${\hat A_t}$ denotes the deviation of action ${a_t}$ from the mean action at the state ${s_t}$. To reduce the deviation, an exponentially-weighting method is employed to obtain the generalized advantage estimation (GAE) \cite{ref34}
    \begin{equation} 
        {\hat A_t} = \sum\limits_{t' = t}^T {{{(\gamma \lambda )}^{t' - t}}\left( {{r_t} + \gamma V_\phi ^\pi ({s_{t + 1}}) - {V'}^\pi ({s_t})} \right)} .
        \label{eq38}  
    \end{equation} 

Since the loss clipping policy has been proven effective in training actor network, the clipped loss can be expressed as
    \begin{equation} 
        {r_t}(\theta ) = \frac{{\pi _\theta ^{}({a_t}|{s_t})}}{{\pi _{\theta old}^{}({a_t}|{s_t})}},
        \label{eq39}  
    \end{equation} 
    \begin{equation} 
        {\mathcal{L}^{CLIP}}(\theta ) = {\mathbb{E}_t}\left[ {\min ({r_t}(\theta ){{\hat A}_t},clip({r_t}(\theta ),1 - \epsilon,1 + \epsilon){{\hat A}_t})} \right],
        \label{eq40}  
    \end{equation}    
where $\epsilon$ is a hyperparameter that controls the range of clipped ${r_t}(\theta )$. Finally, the objective function of the actor network can be written as 
    \begin{equation} 
        {\mathcal{L}_a}(\theta ) = \sum\limits_{m = 1}^M {\left[ {{\mathcal{L}^{CLIP}}({\theta _m}) + \zeta {\mathbb{E}_t}\left[ {\mathcal{H}({\pi _{{\theta _m}}}({a_t}|{s_t}))} \right]} \right]} ,
        \label{eq41}  
    \end{equation}     
where $\mathcal{H}({\pi _{{\theta _m}}}({a_t}|{s_t}))$ is an entropy bonus that encourages exploration and $\zeta$ is a balancing hyperparameter. The critic network is supposed to fit an unknown state value function, while the actor networks should provide a policy to maximize the fitted state value. During the training process, the network parameters are dynamically updated by gradient optimization, which guides the critic and actors to gradually converge toward their optimal objectives. The proposed MAHPPO-AM algorithm is summarized in Algorithm \ref{alg1}. 

 \begin{algorithm}
\caption{MAHPPO-AM Algorithm for UAV Swarm-assisted Integrated Communication and Control Co-design}\label{alg1}
\label{alg1}
\begin{algorithmic}[1]
\STATE Initialize parameters of critic and actor networks as $\phi$ and $\theta_m, m\in\mathcal{M}$;
\STATE Initialize the memory buffer $\bar{M}$, batch size $B$, sample reuse time $\bar{K}$ and learning rate $l_{r}$;
\FOR{episode $=1,\ldots, E_{p}$}
    \STATE Randomly select two obstructed areas; 
    \STATE Randomly generate mobile GUs' locations;
    \STATE Initialize the system state $s_{1}$;
    \FOR{$t=1,\ldots, T$}
        \STATE Sample $a_{t}\sim\pi_{\theta}\left(a_{t}\mid s_{t}\right)$ with action masking;
        \STATE Execute $a_{t}$ to obtain its corresponding reward $r_{t}$ and the next state $s_{t+1}$;
        \STATE Store tuple $\left\langle s_{t}, a_{t}, r_{t}, s_{t+1}\right\rangle$ in the memory buffer $\bar{M}$;
        \STATE Current state $s_{t}\leftarrow s_{t+1}$;
        \IF{$\bar{M}$ is filled}
            \STATE Compute state value in $\bar{M}$ using: \\$V^{\prime\pi}\left(s_{t}\right)=\sum_{t^{\prime}=t}^{T}\gamma^{t^{\prime}-t} r_{t^{\prime}}$;
            \STATE Compute advantage for states in $\bar{M}$ based on \\${\hat A_t} = \sum\limits_{t' = t}^T {{{(\gamma \lambda )}^{t' - t}}\left( {{r_t} + \gamma V_\phi ^\pi ({s_{t + 1}}) - {V'}^\pi ({s_t})} \right)}$;
            \FOR{$i=1,\ldots,\left\lfloor\bar{K}\times(\bar{M}/ B)\right\rfloor$}
                \STATE Sample a mini-batch of $B$ tuples from $\bar{M}$;
                \STATE Compute the loss of critic network: \\$\mathcal{L}_c(\phi)=\left\|V_\phi^\pi\left(s_t\right)-V^{\prime\pi}\left(s_t\right)\right\|_2$;
                \STATE Compute the loss of actor network:\[ \begin{gathered}{\mathcal{L}_a}(\theta ) =  \hfill \\
                \sum\limits_{m = 1}^M {\left[ {{\mathcal{L}^{CLIP}}({\theta _m}) + \zeta {E_t}\left[ {\mathcal{H}({\pi _{{\theta _m}}}({a_t}|{s_t}))} \right]} \right]};  \hfill \\ 
                \end{gathered}   \] 
                \STATE Update critic with $\phi\leftarrow\phi-l_{r}\nabla_{\phi}\mathcal{L}_{c}(\phi)$;
                \FOR{$m=1,\ldots, M$}
                    \STATE Update actor with $\theta_{m}\leftarrow\theta_{m}-l_{r}\nabla_{\theta_{m}}\mathcal{L}_{a}(\theta)$;
                \ENDFOR
            \ENDFOR  
            \STATE Clear memories in $\bar{M}$.
        \ENDIF
    \ENDFOR
\ENDFOR   
\end{algorithmic}  
\end{algorithm}

\section{Simulation Results}
In the 3D coordinate, we focus on an urban area of 1,500 $\times$ 1,500 m$^2$, i.e., $X = 1500$, $Y = 1500$, and $Z = 100$.  The initial positions of the UAVs are fixed. The initial position of user $k$ is randomly and uniformly distributed on the ground. Between any two consecutive time slots, i.e., $t$ and $t+1$, user $k$ moves a distance ${d_k}(t)$ in a direction ${\theta _k}(t)$, where ${d_k}(t)$ and ${\theta _k}(t)$ are uniformly distributed in the ranges of $(0,{d_{\max }})$ and $(0,2\pi )$, respectively. ${d_{\max }}$ denotes the maximum distance a mobile user can move in a time slot. When a mobile user reaches the boundary of the area, it will reverse its direction and continue moving within the coverage area. To better simulate severe obstruction to A2G links caused by ground buildings, two small areas are randomly selected as obstructed areas in the initialization. For default system parameters, Table \ref{table1} summarizes the parameters used in the conducted numerical simulations.    
\begin{table} 
\centering 
\caption{Parameters and Values}
\label{table1}
\renewcommand{\arraystretch}{1.5} 
\begin{tabular}{|>{\raggedright}m{2.5cm}|m{2.5cm}|m{2.5cm}|} 
\hline
\textbf{Parameter} & \textbf{Symbol} & \textbf{Value} \\ \hline
Number of UAVs & $M$ & 3 \\ \hline
Number of GUs & $K$ & 9 \\ \hline
Number of subcarriers & $N$ & 3 \\ \hline
Number of time slots & $T$ & 60 \\ \hline
Duration of each time slot \cite{ref17} & ${T_s}$ & 1s \\ \hline
Altitude of UAVs & $h_m(t)$ & [40m, 100m] \\ \hline
Maximum acceleration of UAVs \cite{3D UAV} & $a_{max}$ & 5m/s$^2$ \\ \hline
Maximum speed of UAVs \cite{3D UAV} & $V_{max}$ & 20m/s \\ \hline
Safe distance between UAVs \cite{ref15} & $d_{min}$ & 25m \\ \hline
Maximum travel distance of GUs & $d_{max}$ & 10m \\ \hline
UAV aerodynamics parameters \cite{ref17} & \makecell[l]{$\rho,A,\delta,c_{S},$\\${S_{FP}},c_{T},c_{f}$} & 1.225kg/m$^3$, 0.79m$^2$, 0.012, 0.1, 0.01m$^2$, 0.3, 0.13 \\ \hline
Weight of the UAV and gravitational acceleration \cite{ref17} & $m', \|\bm{g}\|$ & 2kg, 9.8m/s$^2$ \\ \hline
Maximum power of UAVs \cite{ref15} & $p_{max}$ & 30dBm \\ \hline
Carrier frequency \cite{ref20} & $f_c$ & 2GHz \\ \hline
Noise density \cite{ref15} & $\sigma^2$ & -107dBm \\ \hline
MAHPPO-AM algorithm parameters \cite{MAHPPO} &\makecell[l]{$l_r,B,\bar K,\bar M,E_p,$\\$\gamma,\lambda,\epsilon,\zeta$} & 0.0001, 256, 20, 2048, 2000, 0.95, 0.95, 0.2, 0.001 \\ \hline
\end{tabular}
\end{table}

In the proposed MAHPPO-AM algorithm, each actor network consists of fully connected layers, where the first two layers are the shared layers containing 256 and 128 neurons, respectively. Also, each output branch includes 2 layers, and the first layer contains 64 neurons, as well as the structure of the last layer is determined by the type and dimension of its corresponding actions. The critic network consists of 4 fully connected layers containing 256, 128, 64, and 1 neuron, respectively.

For comparison, we adopt the following two schemes:
\begin{itemize}\item[$\bullet$]Hybrid Proximal Policy Optimization (HPPO) \cite{ref35}: As a single-agent algorithm, all UAVs share the same actor network to generate the corresponding actions simultaneously.  
\item[$\bullet$]Exploration: It has the same framework as the MAHPPO-AM algorithm. Still, the agents only perform the current policy to interact with the environment and collect experience without gradient updates on the policy parameters. 
\end{itemize}

\begin{figure*}
\begin{minipage}[t]{0.5\textwidth}
\centerline{\includegraphics[width=3.2in]{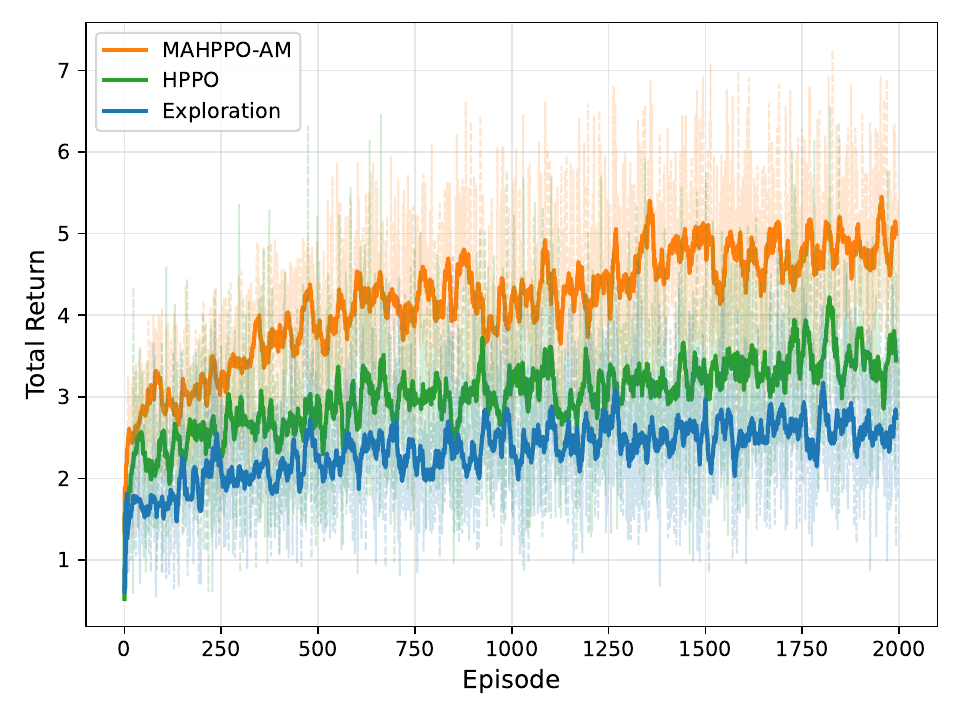}}
\caption{Training process of the MAHPPO-AM algorithm and the two
\ 
\newline baseline methods.}
\label{fig3}
\end{minipage}%
\begin{minipage}[t]{0.5\textwidth}
\centerline{\includegraphics[width=3.5in]{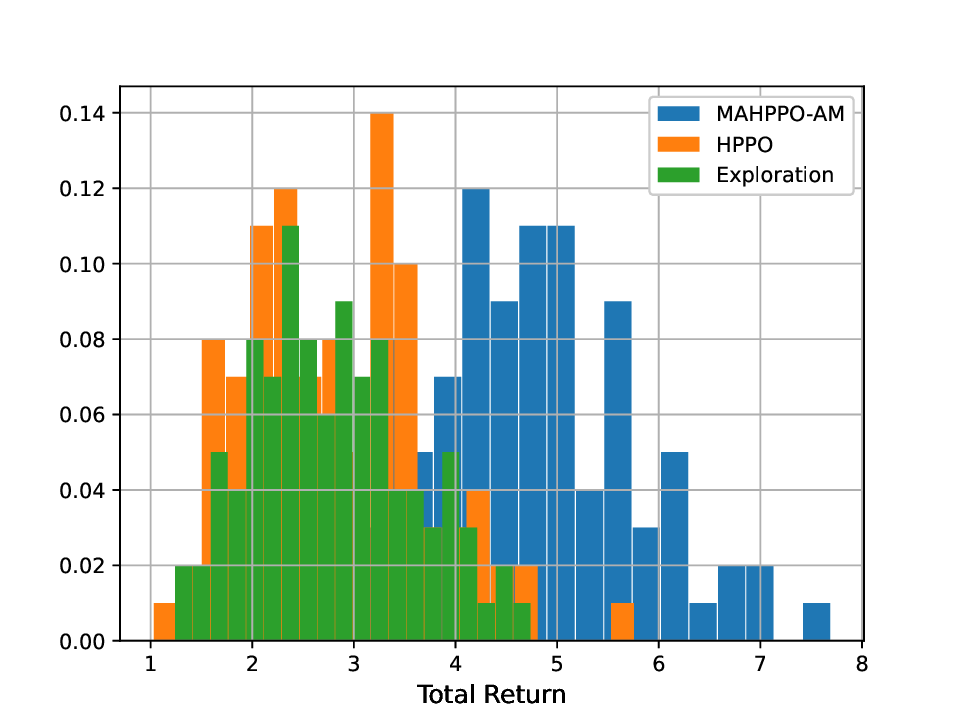}}
\caption{Histogram of total return for the MAHPPO-AM algorithm and the two baseline methods.}
\label{fig4}
\end{minipage}
\begin{minipage}[t]{0.5\textwidth}
\centerline{\includegraphics[width=4in]{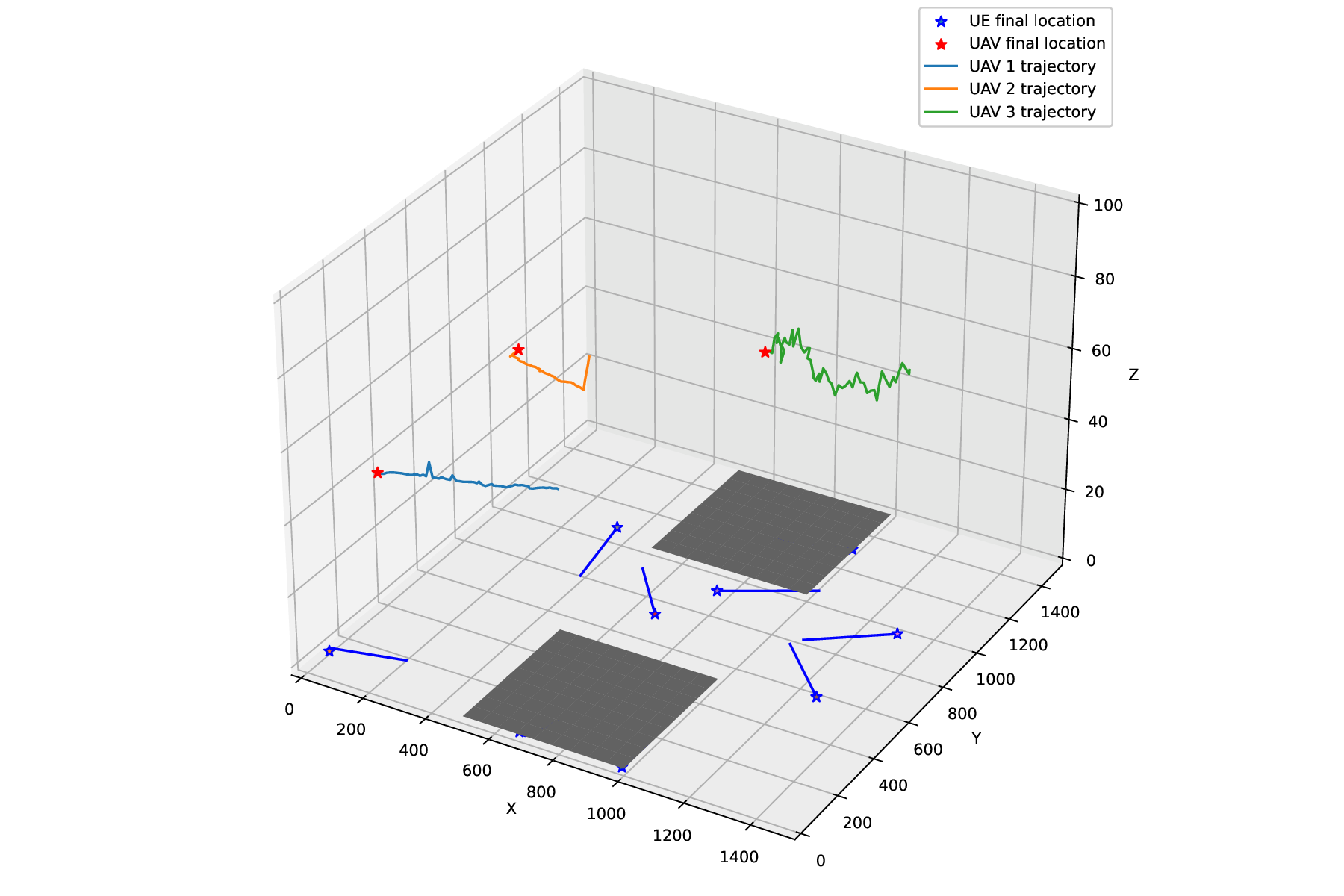}}
\caption{3D Trajectory of the UAV swarm.}
\label{fig5}
\end{minipage}
\begin{minipage}[t]{0.5\textwidth}
\centerline{\includegraphics[width=3.5in]{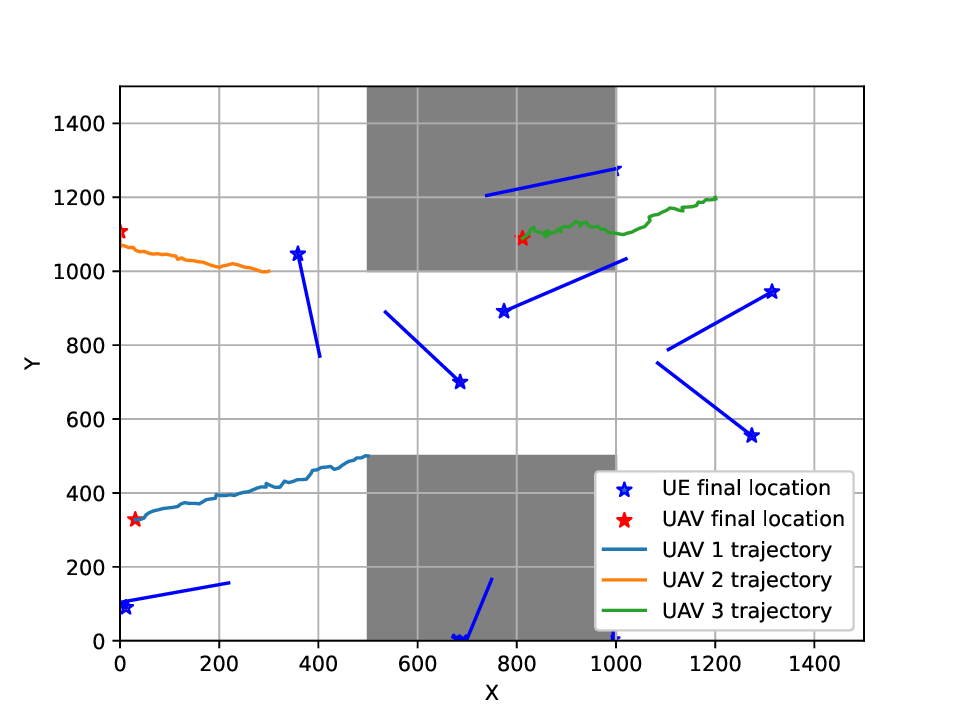}}
\caption{2D Trajectory of the UAV swarm.}
\label{fig6}
\end{minipage}%
\end{figure*}
Fig. \ref{fig3} shows the training process of the MAHPPO-AM algorithm and the two baseline methods. The curves are smoothed by taking the average of the five nearest values at each point. The proposed MAHPPO-AM method achieves optimal convergence performance over 2,000 training episodes, with its total return steadily increasing from 3.5 to around 5. This trend suggests that the proposed algorithm is effective in learning a policy that balances the mobile users’ fair communication rates and the UAVs’ energy consumption. In contrast, HPPO eventually converges to a total return between 3 and 4, which reflects the limitations of a single agent when dealing with UAV swarm collaboration. With total return converging between 2 and 3, the Exploration method demonstrates the ineffectiveness of just collecting data without updating the policy parameters. Fig. \ref{fig4} shows the histograms obtained over 100-time simulation for the three methods. According to the statistics, the average return for the MAHPPO-AM, HPPO, and Exploration is 4.60, 2.82, and 2.80, respectively. These results show that the MAHPPO-AM method performs relatively well in training and simulation.

Fig. \ref{fig5} and Fig. \ref{fig6} depict the 3D trajectory and 2D trajectory of the UAV swarm obtained by the MAHPPO-AM method, respectively. And the gray area is the obstructed area. The 3D trajectory provides a comprehensive view of the UAV swarm’s spatial movement and altitude variation, while the 2D trajectory clearly illustrates the swarm’s coverage of mobile users and the spatial coordination among UAVs. By comparing the two figures, UAV 1 and UAV 2 can optimize their flight trajectory by minimizing unnecessary travel distance and altitude adjustments while maintaining communication quality, so as to reduce energy consumption. The trajectories of the UAV swarm reflect that the MAHPPO-AM method is able to intelligently allocate the coverage area of UAVs based on the spatial distribution of mobile users. Moreover, the UAV swarm is capable of providing fair service for mobile users with lower energy consumption, while avoiding situation in which certain users cannot be served due to their presence in obstructed areas.     

\begin{figure*}
\begin{minipage}[t]{0.5\textwidth}
\centerline{\includegraphics[width=3.5in]{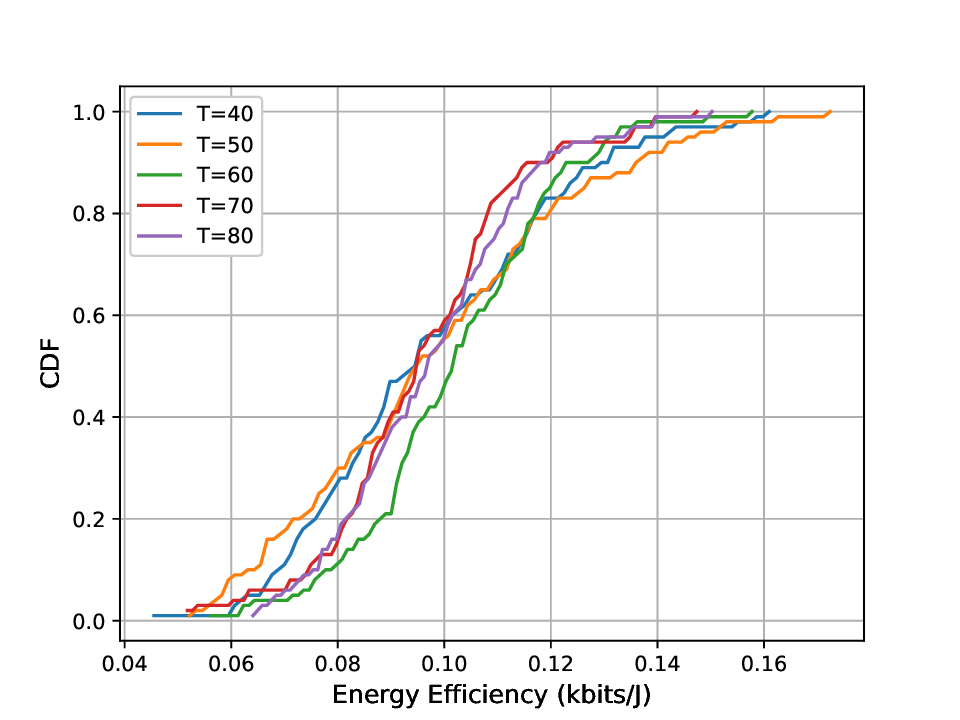}}
\caption{Energy efficiency comparison for different $T$ steps.}
\label{fig7}
\end{minipage}%
\begin{minipage}[t]{0.5\textwidth}
\centerline{\includegraphics[width=3.5in]{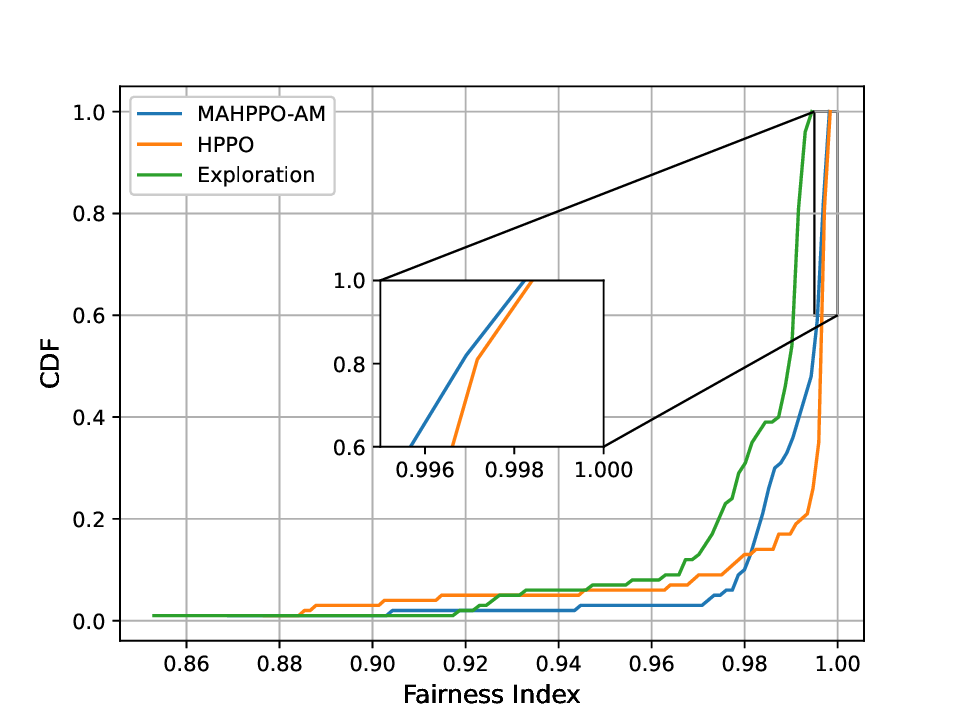}}
\caption{Fairness index comparison for the MAHPPO-AM algorithm and the two baseline methods.}
\label{fig8}
\end{minipage}%
\end{figure*}

\begin{table}
\centering
\caption{ENERGY EFFICIENCY COMPARISON}
\label{table2}
\renewcommand{\arraystretch}{1.5} 
\begin{tabular}{|c|c|c|c|c|c|}
\hline
Time Slots & T=40s & T=50s & T=60s & T=70s & T=80s \\ \hline
\makecell[c]{Energy Efficiency \\ (kbits/J)} & 0.0978 & 0.0984 & 0.1024 & 0.0961 & 0.0979 \\ \hline
\end{tabular}
\end{table}
Based on the proposed MAHPPO-AM method, Fig. \ref{fig7} compares the cumulative distribution function (CDF) of the average long-term energy efficiency over 100-time simulation for different $T$ steps. As summarized in Table \ref{table2}, the system achieves the optimal energy efficiency when $T = 60$s. It is worth noting that the energy efficiency instead decreases at $T = 70$s, which indicates that the energy efficiency is not positively correlated with $T$. Overall, increasing $T$ tends to enhance system energy efficiency, but $T = 60$s appears to be a threshold beyond which further increases lead to performance degradation. Therefore, we set $T = 60$s in our simulation experiments.


Fig. \ref{fig8} illustrates the CDF of the fairness index over 100-time simulation, where the average fairness index for the MAHPPO-AM, HPPO, and Exploration is 0.9888, 0.9887, and 0.9808, respectively. As shown in the figure, the communication fairness of the three methods performs well, largely due to the effectiveness of the action masking applied during training. Although the average fairness index of MAHPPO-AM and HPPO methods is nearly identical, MAHPPO-AM performs more prominently in the high-fairness interval. Specifically, when the fairness index exceeds 0.99, the probability of MAHPPO-AM is greater than that of the HPPO method. In contrast, the fairness index of Exploration method is significantly inferior to the proposed approach. These results demonstrate that the proposed MAHPPO-AM method provides more reliable communication services in high fairness demand scenarios.

\begin{figure}[!t]
\centering
\includegraphics[width=3.5in]{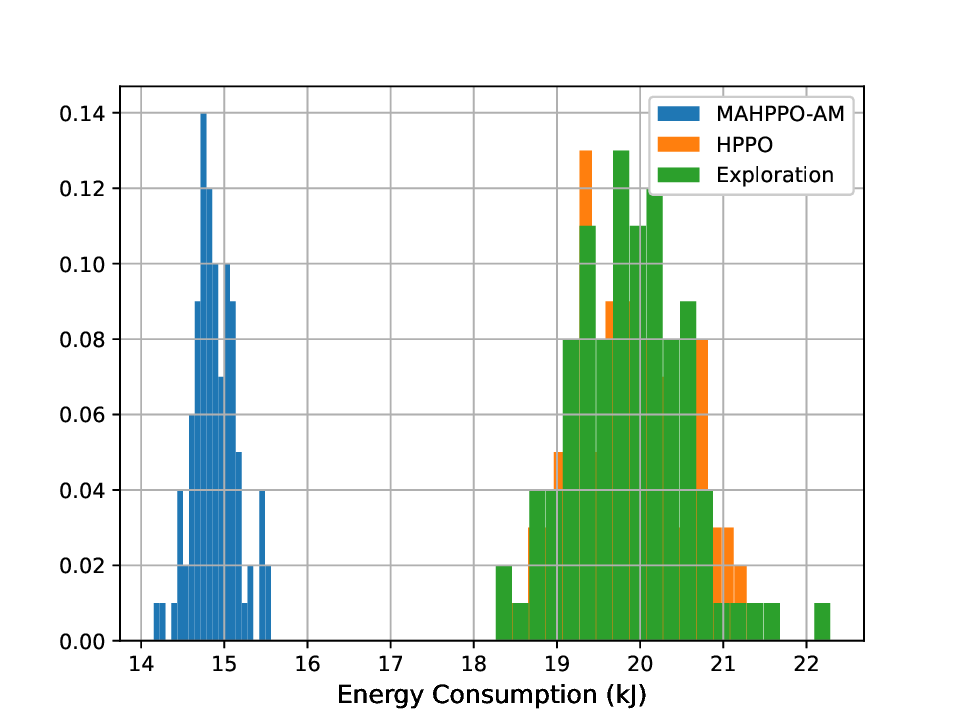}
\caption{Histogram of energy consumption for the MAHPPO-AM algorithm and the two baseline methods.}
\label{fig9}
\end{figure}
Fig. \ref{fig9} displays the histograms of the energy consumption over 100-time simulation obtained by different methods, where the average energy consumption for the MAHPPO-AM, HPPO, and Exploration is 14.88kJ, 19.86kJ, and 19.88kJ, respectively. As observed, the MAHPPO-AM method reduces average energy consumption by approximately 25$\% $ compared to the baseline methods. Meanwhile, the energy consumption distribution of MAHPPO-AM shows a sharp and narrow peak, indicating that its energy consumption is also more stable. The above results demonstrate that the MAHPPO-AM method executes a more energy-efficient  policy.   

\section{Conclusion}

This paper investigates a UAV swarm-assisted integrated communication and control co-design mechanism to enhance communication fairness and quality of service in complex geographic environment. Considering the energy constraints of UAV swarms, we formulate an objective function that characterizes the trade-off between the fairness-constrained communication rates for ground users and UAVs’ energy consumption. We model the UAV swarm’s resource allocation and 3D trajectory control as a Markov decision process and propose a multi-agent RL framework as a solution, which facilitates decision-making on collaborative actions among UAVs in real time. To address hybrid action space challenges in UAV swarms, we propose a novel MAHPPO-AM algorithm that guarantees hard constraints in high-dimensional action spaces via action masking. The simulation results show that our approach achieves a fairness index of 0.99 while reducing energy consumption by up to 25$\%$ compared to baseline methods.




 




\vfill


\begin{thebibliography}{1}
\bibliographystyle{IEEEtran}

\bibitem{ref1}
Zeng Y, Wu Q, Zhang R, ``Accessing From the Sky: A Tutorial on UAV Communications for 5G and Beyond,"{\it{Proceedings of the IEEE}}. 2019, 107(12): 2327–2375.

\bibitem{ref2}
Luo Q, Luan T H, Shi W, Fan P, ``Deep Reinforcement Learning Based Computation Offloading and Trajectory Planning for Multi-UAV Cooperative Target Search",{\it{ IEEE Journal on Selected Areas in Communications}}. 2023, 41(2): 504–520.

\bibitem{ref3}
Sun W-B, Zhao L, Yang X, Wang L, Meng W-X, ``Joint Topology Reconstruction and Resource Allocation for UAV-IoT Networks,"{\it{IEEE Internet of Things Journal}}, 2024: 1–1.

\bibitem{ref4}
Ning Z, Yang Y, Wang X, Song Q, Guo L, Jamalipour A, ``Multi-Agent Deep Reinforcement Learning Based UAV Trajectory Optimization for Differentiated Services,"{\it{IEEE Transactions on Mobile Computing}}, 2023: 1–17.

\bibitem{ref5}
A. Bera, S. Misra, C. Chatterjee and S. Mao, ``CEDAN: Cost-Effective Data Aggregation for UAV-Enabled IoT Networks," {\it{IEEE Transactions on Mobile Computing,}} vol. 22, no. 9, pp. 5053-5063, 1 Sept. 2023, doi: 10.1109/TMC.2022.3172444. 

\bibitem{ref6}
Xing R, Su Z, Luan T H, Xu Q, Wang Y, Li R, ``UAVs-Aided Delay-Tolerant Blockchain Secure Offline Transactions in Post-Disaster Vehicular Networks," \textit{IEEE Transactions on Vehicular Technology},2022, 71(11): 12030–12043.

\bibitem{ref7}
Gaydamaka A, Samuylov A, Moltchanov D, Ashraf M, Tan B, Koucheryavy Y, ``Dynamic Topology Organization and Maintenance Algorithms for Autonomous UAV Swarms," in \textit{IEEE Transactions on Mobile Computing}, 2023: 1–17.

\bibitem{ref8}
Wu Q, Xu J, Zeng Y, Ng D W K, Al-Dhahir N, Schober R, Swindlehurst A L, ``A Comprehensive Overview on 5G-and-Beyond Networks With UAVs: From Communications to Sensing and Intelligence," \textit{IEEE Journal on Selected Areas in Communications}, 2021, 39(10): 2912–2945.  

\bibitem{ref9}
Zhou Z, Zhuang Y, Li H, Huang S, Yang S, Guo P, Zhong L, Yuan Z, Xu C. MR-FFL, ``A Stratified Community-Based Mutual Reliability Framework for Fairness-Aware Federated Learning in Heterogeneous UAV Networks,"in \textit{IEEE Internet of Things Journal}, 2024: 1–1.

\bibitem{ref10} 
C. Qiu, Z. Wei, X. Yuan, Z. Feng, and P. Zhang, ``Multiple UAV-mounted base station placement and user association with joint fronthaul and backhaul optimization,'' \textit{IEEE Trans. Commun.}, vol. 68, no. 9, pp. 5864--5877, Sep. 2020.

\bibitem{ref11} 
C. Shen, T.-H. Chang, J. Gong, Y. Zeng, and R. Zhang, ``Multi-UAV interference coordination via joint trajectory and power control,'' \textit{IEEE Trans. Signal Process.}, vol. 68, pp. 843--858, 2020.

\bibitem{ref12} 
X. Liu, B. Lai, B. Lin, and V. C. M. Leung, ``Joint communication and trajectory optimization for multi-UAV enabled mobile Internet of Vehicles,'' \textit{IEEE Trans. Intell. Transp. Syst.}, vol. 23, no. 9, pp. 15354--15366, Sep. 2022.

\bibitem{ref13} 
L. Zhu, J. Zhang, Z. Xiao, X.-G. Xia, and R. Zhang, ``Multi-UAV aided millimeter-wave networks: Positioning, clustering, and beamforming,'' \textit{IEEE Trans. Wireless Commun.}, vol. 21, no. 7, pp. 4637--4653, Jul. 2022.

\bibitem{ref14} 
R. Zhang, Y. Zhang, R. Tang, H. Zhao, Q. Xiao and C. Wang, ``A Joint UAV Trajectory, User Association, and Beamforming Design Strategy for Multi-UAV-Assisted ISAC Systems,'' \textit{IEEE Internet of Things Journal}, vol. 11, no. 18, pp. 29360--29374, 15 Sept. 2024, doi: 10.1109/JIOT.2024.3430390.

\bibitem{ref15} 
P. Yi, L. Zhu, Z. Xiao, R. Zhang, Z. Han and X.-G. Xia, ``3-D Positioning and Resource Allocation for Multi-UAV Base Stations Under Blockage-Aware Channel Model,'' \textit{IEEE Transactions on Wireless Communications}, vol. 23, no. 3, pp. 2453--2468, March 2024, doi: 10.1109/TWC.2023.3300332.


\bibitem{ref18} 
X. Liu, M. Chen, Y. Liu, Y. Chen, S. Cui and L. Hanzo, ``Artificial Intelligence Aided Next-Generation Networks Relying on UAVs,'' \textit{IEEE Wireless Communications}, vol. 28, no. 1, pp. 120--127, February 2021, doi: 10.1109/MWC.001.2000174.

\bibitem{ref19} 
S. Yin and F. R. Yu, ``Resource Allocation and Trajectory Design in UAV-Aided Cellular Networks Based on Multiagent Reinforcement Learning,'' \textit{IEEE Internet of Things Journal}, vol. 9, no. 4, pp. 2933--2943, 15 Feb. 2022, doi: 10.1109/JIOT.2021.3094651.

\bibitem{ref20} 
R. Zhong, X. Liu, Y. Liu and Y. Chen, ``Multi-Agent Reinforcement Learning in NOMA-Aided UAV Networks for Cellular Offloading,'' \textit{IEEE Transactions on Wireless Communications}, vol. 21, no. 3, pp. 1498--1512, March 2022, doi: 10.1109/TWC.2021.3104633.

\bibitem{ref21} 
W. J. Yun et al., ``Cooperative Multiagent Deep Reinforcement Learning for Reliable Surveillance via Autonomous Multi-UAV Control,'' \textit{IEEE Transactions on Industrial Informatics}, vol. 18, no. 10, pp. 7086--7096, Oct. 2022, doi: 10.1109/TII.2022.3143175.

\bibitem{ref22} 
X. Wang, M. Yi, J. Liu, Y. Zhang, M. Wang, and B. Bai, ``Cooperative data collection with multiple UAVs for information freshness in the Internet of Things,'' \textit{IEEE Trans. Commun.}, vol. 71, no. 5, pp. 2740--2755, May 2023.

\bibitem{ref23} 
Q. Gao, R. Zhong, H. Shin and Y. Liu, ``MARL-Based UAV Trajectory and Beamforming Optimization for ISAC System,'' in \textit{in IEEE Internet of Things Journal}, vol. 11, no. 24, pp. 40492-40505, 15 Dec.15, 2024, doi: 10.1109/JIOT.2024.3453195.

\bibitem{ref24} 
S. Xu, X. Zhang, C. Li, D. Wang, and L. Yang, ``Deep reinforcement learning approach for joint trajectory design in multi-UAV IoT networks,'' \textit{IEEE Trans. Veh. Technol.}, vol. 71, no. 3, pp. 3389--3394, Mar. 2022.

\bibitem{ref25} 
K. Li, W. Ni, Y. Emami, and F. Dressler, ``Data-driven flight control of Internet-of-Drones for sensor data aggregation using multi-agent deep reinforcement learning,'' \textit{IEEE Wireless Commun.}, vol. 29, no. 4, pp. 18--23, Aug. 2022.

\bibitem{ref26} 
J. Ji, K. Zhu, and L. Cai, ``Trajectory and communication design for cache-enabled UAVs in cellular networks: A deep reinforcement learning approach,'' \textit{IEEE Trans. Mobile Comput.}, vol. 22, no. 10, pp. 6190--6204, Oct. 2023.

\bibitem{ref27} 
Z. Bai, J. Shi, Z. Li, M. Li and X. Liao, ``An MA-HPPO Approach for Multi-UAV Data Collection,'' \textit{IEEE Transactions on Wireless Communications}, vol. 23, no. 12, pp. 17974--17986, Dec. 2024, doi: 10.1109/TWC.2024.3458194.

\bibitem{ref28} 
``Technical Specification Group Radio Access Network; Study on Enhanced LTE Support for Aerial Vehicles,'' 36.777, 3GPP, 2018, version 15.0.0.

\bibitem{TWC-2018} 
J. Lyu, Y. Zeng, and R. Zhang, ``UAV-aided offloading for cellular hotspot,'' \textit{IEEE Trans. Wireless Commun.}, vol. 17, no. 6, pp. 3988–4001, Jun. 2018.

\bibitem{ref29} 
C. H. Liu, Z. Chen, J. Tang, J. Xu, and C. Piao, ``Energy-efficient UAV control for effective and fair communication coverage: A deep reinforcement learning approach,'' \textit{IEEE J. Sel. Areas Commun.}, vol. 36, no. 9, pp. 2059--2070, Sep. 2018.

\bibitem{ref30} 
R. K. Jain, D.-M. W. Chiu, and W. R. Hawe, ``A quantitative measure of fairness and discrimination,'' Eastern Res. Lab., Digit. Equip. Corp., Hudson, MA, USA, 1984.

\bibitem{ref31} 
Y. Zeng and R. Zhang, ``Energy-efficient UAV communication with trajectory optimization,'' \textit{IEEE Trans. Wireless Commun.}, vol. 16, no. 6, pp. 3747--3760, Jun. 2017.

\bibitem{ref32} 
S. Huang and S. Ontañón, ``A Closer Look at Invalid Action Masking in Policy Gradient Algorithms,'' \textit{FLAIRS}, vol. 35, May 2022.

\bibitem{ref33} 
J. Schulman, S. Levine, P. Abbeel, M. I. Jordan, and P. Moritz, ``Trust region policy optimization,'' in \textit{Proc. Int. Conf. Mach. Learn. (ICML)}, Lille, France, Jul. 2015, pp. 1889--1897.

\bibitem{ref17} 
X. Dai, B. Duo, X. Yuan and M. D. Renzo, ``Energy-Efficient UAV Communications in the Presence of Wind: 3D Modeling and Trajectory Design,'' \textit{IEEE Transactions on Wireless Communications}, vol. 23, no. 3, pp. 1840--1854, March 2024, doi: 10.1109/TWC.2023.3292290.

\bibitem{ref34} 
J. Schulman, P. Moritz, S. Levine, M. I. Jordan, and P. Abbeel, ``High-dimensional continuous control using generalized advantage estimation,'' in \textit{Proc. Int. Conf. Learn. Represent. (ICLR)}, San Juan, Puerto Rico, May 2016.

\bibitem{3D UAV} 
R. Ding, F. Gao and X. S. Shen, ``3D UAV Trajectory Design and Frequency Band Allocation for Energy-Efficient and Fair Communication: A Deep Reinforcement Learning Approach,'' in \textit{IEEE Transactions on Wireless Communications}, vol. 19, no. 12, pp. 7796-7809, Dec. 2020, doi: 10.1109/TWC.2020.3016024.

\bibitem{MAHPPO} 
Z. Hao, G. Xu, Y. Luo, H. Hu, J. An and S. Mao,``Multi-Agent Collaborative Inference via DNN Decoupling: Intermediate Feature Compression and Edge Learning,'' in \textit{IEEE Transactions on Mobile Computing}, vol. 22, no. 10, pp. 6041-6055, 1 Oct. 2023, doi: 10.1109/TMC.2022.3183098.

\bibitem{ref35} 
Z. Fan, R. Su, W. Zhang et al., ``Hybrid Actor-Critic Reinforcement Learning in Parameterized Action Space,'' in \textit{Proc. Int. Joint Conf. Artif. Intell. (IJCAI)}, 2019.

 
\end{thebibliography}
\end{document}